\documentclass[letterpaper]{article} %
\usepackage[submission]{aaai24}  %
\usepackage{times}  %
\usepackage{helvet}  %
\usepackage{courier}  %
\usepackage[hyphens]{url}  %
\usepackage{graphicx} %
\usepackage{natbib}  %
\usepackage{caption} %
\usepackage{algorithm}
\usepackage{algorithmic}

\usepackage{newfloat}
\usepackage{listings}
\DeclareCaptionStyle{ruled}{labelfont=normalfont,labelsep=colon,strut=off} %
\floatstyle{ruled}
\newfloat{listing}{tb}{lst}{}
\floatname{listing}{Listing}
\usepackage{graphicx}
\usepackage{multirow}
\usepackage{enumitem}
\usepackage[at]{easylist}%
\usepackage{booktabs}
\usepackage{amssymb}
\usepackage{kotex}

\title{Beyond Weaponization: NLP Security for Medium and \\ Lower-Resourced Languages in Their Own Right}

\author {
    Heather Lent
}
\affiliations{
    Department of Computer Science\\
    Aalborg University\\
    hcle@cs.aau.dk
}

\begin{document}

\maketitle

\begin{abstract}
Despite mounting evidence that multilinguality can be easily weaponized against language models (LMs), works across NLP Security remain overwhelmingly English-centric. 
In terms of securing LMs, the NLP norm of ``English first'' collides with standard procedure in cybersecurity, whereby practitioners are expected to anticipate and prepare for \textit{worst-case} outcomes. 
To mitigate worst-case outcomes in NLP Security, researchers must be willing to engage with the weakest links in LM security: lower-resourced languages. 
Accordingly, this work examines the security of LMs for lower- and medium-resourced languages. 
We extend existing adversarial attacks for up to 70 languages to evaluate the security of monolingual and multilingual LMs for these languages.
Through our analysis, we find that monolingual models are often too small in total number of parameters to ensure sound security, and that while multilinguality is helpful, it does not always guarantee improved security either.
Ultimately, these findings highlight important considerations for 
more secure deployment of LMs, for communities of lower-resourced languages.
\end{abstract}

\section{Introduction}
The mass proliferation of language models (LMs) is unraveling the longstanding asymmetries between higher, medium, and lower-resourced languages (HRLs, MRLs, LRLs) in NLP. 
Previously, the lack of strong support for MRLs and LRLs affected speakers of those languages \textit{alone}; now, with millions of users interacting with LMs every day, the lack of robustness for LRLs in LMs stands to impact \textit{everyone}, as LRLs are being directly weaponized against LM security \cite{ 10.1145/3630106.3658546, yong2024lowresource, lent2025nlpsecurityethicswild}.
Works across NLP Security are sounding the alarm on the potential for malicious actors to exploit multilinguality to launch wide variety of effective attacks against LMs \cite{chen-etal-2024-text, he2024transferring, wang-etal-2024-backdoor, wang-etal-2024-languages}, even where reasonable defenses exist for English.
For example, \citet{yong2024lowresource} demonstrate that LRLs can be leveraged to successfully jailbreak GPT4's safety filter in 79\% of instances, in comparison to English's $<$1\% attack success rate. 
As long as LRLs and MRLs can be weaponized to punch through LM defenses, the broader public will remain vulnerable to a sweeping array of security threats by bad actors, from phishing to misinformation campaigns, regardless of the language(s) they speak.

\begin{figure}[t]
    \centering
    \includegraphics[width=0.9\columnwidth]{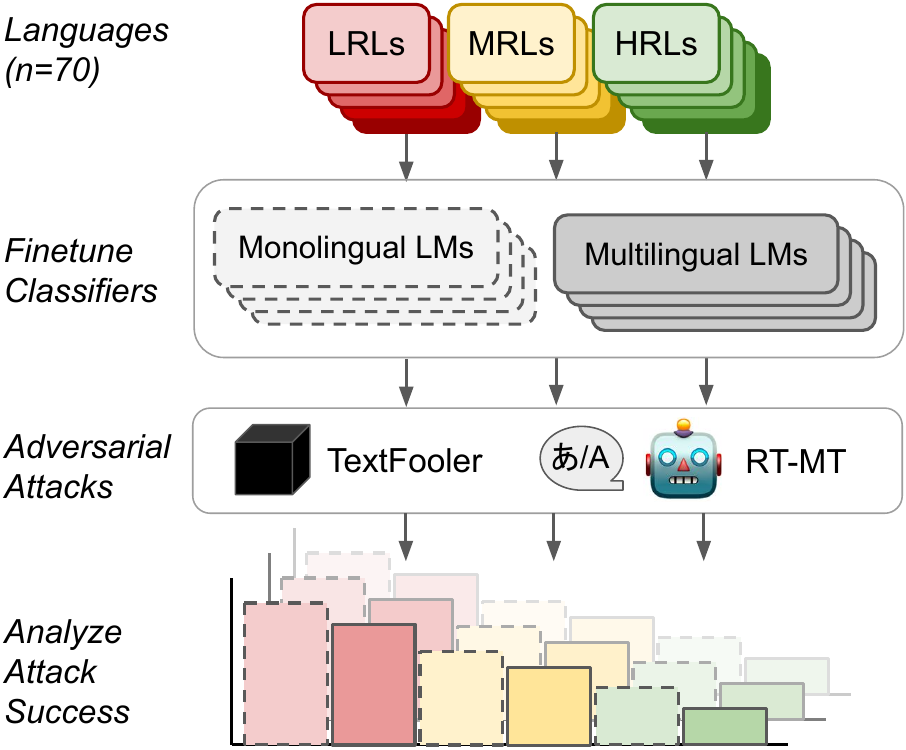}
    \caption{
    We investigate LM security for 70 lower, medium, and higher-resourced languages (LRLs, MRLs, and HRLs) across a variety of monolingual and multilingual LMs.
    We extend the popular baseline black-box adversarial attack TextFooler \cite{jin2020bertreallyrobuststrong} to a multilingual setting and compare against a baseline attack from round-trip machine translation (RT-MT).  
    }
    \label{fig:fig1}
\end{figure}

The burden of these looming security threats is not shared equally, however. Communities speaking MRLs and LRLs may have the most to lose from insufficient protections for LMs. 
Consider the well-established connection between data availability in NLP and global inequality \cite{blasi-etal-2022-systematic, ranathunga-de-silva-2022-languages};  insufficient LM security for LRLs could exacerbate existing risks to potentially disadvantaged communities \cite{lent2025nlpsecurityethicswild}.
Alarmingly, such cases are already being observed in the wild.  \citet{10.1145/3630106.3658546} document how bad actors are exploiting poor LM performance over Amharic to push harmful content onto vulnerable groups in Ethiopia and beyond.  
Accordingly, these outstanding threats and their consequences thrust an overwhelming burden on governments, organizations, and businesses wishing to use LMs to serve speakers of minority languages and language varieties \cite{foo-khoo-2025-lionguard, lim2025safemarginsgeneralapproach}.

Amidst the urgency of this present situation, no previous works in NLP Security have carefully examined LRLs or MRLs, likely because there are many challenges inherent to extending existing NLP Security methodologies to these languages. For example, many common attack generation techniques require \textit{inter alia} dictionaries \cite{ren-etal-2019-generating} or large quantities of data to train paraphrasing models \cite{lei-etal-2022-phrase}, which are simply not available for the vast majority of languages languages \cite{joshi-etal-2020-state}. 
Despite these complications, this work sets out to examine LM security for LRLs and MRLs, in the context of adversarial attacks (Figure~\ref{fig:fig1}).
In doing so, we make the following contributions: 

\begin{itemize}[noitemsep,topsep=1pt]
    \item While other works \textit{weaponize} LRLs to red-team LMs, we present the first study on LM security for LRLs and MRLs, in their own right. 
    \item  We complete security testing for up to 70 languages and language varieties. Of these, 29 are LRLs, 33 are MRLs, and 8 are HRLs. The majority of these languages are evaluated in an NLP Security context for the first time. 
    \item  We extend the popular synonym-based adversarial attack TextFooler \cite{jin2020bertreallyrobuststrong} to a multilingual setting, thus building a bridge between the large body English-only NLP Security and the rest of the world's languages. 
    \item We include a round-trip machine translation (MT) attack baseline, ensuring our work can is also comparable with other works in multilingual NLP Security, relying heavily on MT. 
    \item  We experiment with both monolingual and multilingual LMs for every language, providing analyses to assess the current threat-level for MRLs and LRLs. 
\end{itemize}

Finally, given the heightened sensitivity necessary for working with LRLs \cite{bird-2022-local, lent-etal-2022-creole, mager-etal-2023-ethical}, our experiments are designed with ethical best practices in mind. 
While this work grapples with the challenges of extending existing attack frameworks multilingually, we hope it will lay the ground work for the broader field to start developing defense methods for LMs, specifically for LRLs and MRLs.

\section{Related Work}

\paragraph{Adversarial Attacks For English and Beyond.} 
In an adversarial attack, an imagined bad actor passes minimally corrupted inputs to a model, intended to circumvent a model's expected behavior (\textit{e.g.}, misclassification). 
Unlike in computer vision, where additional noise is typically imperceptible to humans \cite{Akhtar2018ThreatOA}, adversarial attacks over natural language are expected retain a high degree of fluency to remain non-suspicious \cite{dyrmishi-etal-2023-humans}. 
For this reason, the majority of \textit{black-box} adversarial attacks (whereby the attacker does not have full access or knowledge of the victim model) in NLP have thus far been largely focused on introducing \textbf{phrase}-level (\textit{e.g.}, \citet{qi-etal-2021-mind, lei-etal-2022-phrase}), \textbf{word}-level (\textit{e.g.}, \citet{alzantot-etal-2018-generating, ren-etal-2019-generating, jin2020bertreallyrobuststrong, li-etal-2020-bert-attack, zang-etal-2020-word, fang-etal-2023-modeling, zhou-etal-2024-evaluating-validity}), and \textbf{character}-level perturbations (\textit{e.g.}, \citet{JiDeepWordBug18, pruthi-etal-2019-combating, Abad2024Charmer}) to inputs.  

Current methods for adversarial attack generation cannot be trivially extended to include LRLs and MRLs, however. Approaches like phrase-level perturbations \cite{lei-etal-2022-phrase} and style-transfer \cite{qi-etal-2021-mind} rely on sizable corpora from diverse domains, which are outside the realm of possibility for the majority of languages. 
And while newer methods rely on LMs to generate adversarial samples \cite{jha2024llmstingerjailbreakingllmsusing}, performance of these models for LRLs is known to be poor \cite{chang-etal-2024-multilinguality}.
Meanwhile, word-level attacks require resources such as dictionaries or lexical databases (\textit{e.g.}, \citet{jin2020bertreallyrobuststrong} use WordNet) for looking up synonyms, as well as POS or NER taggers \cite{ren-etal-2019-generating} for ensuring synonym substitutions remain grammatical. 
For these reasons, word-base attacks are typically crafted for a specific language in mind, and its available resources, rather than scaled multilingually.
For example, word-level adversarial attacks for Chinese have instead leveraged language-specific resources like OpenHownet \cite{morris2020textattack}. 
Similarly for Arabic, \citet{alshahrani-etal-2024-arabic} rely on monolingual AraBERT \cite{antoun-etal-2020-arabert} for synonym lookup.

Finally, character-level attacks attempt to mimic natural-occurring typos, usually for the Latin script \cite{JiDeepWordBug18, Abad2024Charmer}; these methods cannot be trivially extended across all scripts, while retaining stealthiness, however. For instance,  while random character deletion or replacement can yield natural-looking typos for English, this approach can change the meaning of the input for Chinese. For Chinese, a more natural typo would result from incorrect tone. Accordingly, character-level attacks have been designed specifically for Chinese, to substitute homophones and morphonyms (\textit{i.e.}, characters that share similar glyph structures) \cite{morris2020textattack}. 
Similarly, expected typos in Arabic are more likely to concern diacritics \cite{Alslman10989415}, while 
deleting or swapping word-start or word-end characters could result in illegal character combinations \cite{Nakhleh2024CharacterlevelAA}. 
As such, current character-based attacks against new languages are typically custom tailored to the script at hand; such was done for the Tibetan script by \citet{cao-etal-2023-pay-attention}.

Ultimately, current adversarial attacks methods only readily apply to English (or other HRLs with Latin script), thus stymieing efforts to investigate LM Security for diverse MRLs and LRLs.

\paragraph{Multilingual NLP Security and Safety.}

As multilingual models already lag in performance compared to their monolingual counterparts \cite{pfeiffer-etal-2022-lifting}, it is no surprise the LM security also falls behind. Recent works highlight this disparity in the security context, employing a variety of attacks against multilingual LMs, from jailbreaks \cite{yong2024lowresource, wang-etal-2024-languages, Kumar2024AdversarialTO} and backdoor attacks \cite{he2024transferring, wang-etal-2024-backdoor} to embedding inversion attacks \cite{chen-etal-2024-text, Chen_Biswas_Lent_Bjerva_2025}.

The direct weaponization of multilinguality is a common technique across works within this space. For example, \citet{yong2024lowresource} use simple MT from English into LRLs to bypass GPT4's safety filter. 
In this same context of MT-based jailbreaking, analysis from \citet{deng2024multilingualjailbreakchallengeslarge} underscores that, while LMs are already unintentionally less safe for LRLs, deliberate attacks significantly exacerbate the existing threat level.  
Beyond translating whole inputs, \citet{upadhayay-behzadan-2024-sandwich} find that combining a number of prompts in a mixture of languages also successfully break safety mechanisms, and that these mechanisms rely substantially more on English than other languages.  
Even more fine-grained, \citet{yoo2024codeswitchingredteamingllmevaluation} demonstrate how synthetic, intra-phrase code-mixing is an effective method for red-teaming LMs. 
Whether language-level, prompt-level, or phrase-level, MT serves as an essential engine for attack generation or evaluation in multilingual NLP Security \cite{wang-etal-2024-languages}. 
However, the persistent reliance of MT risks under-estimating the severity of current security vulnerabilities.
Translated texts are demonstrably more simple than non-translated counterparts \cite{lembersky-etal-2012-language}, typically leading to significantly inflated performance assessments across a variety of tasks \cite{lubli2018machinetranslationachievedhuman, toral-etal-2018-attaining, artetxe-etal-2020-translation, graham-etal-2020-statistical}. 
In turn, over-reliance of MT across works in multilingual NLP Security potentially risk underestimating the severity of existing threats for MRLs and LRLs.

\begin{figure*}[t]
    \centering
    \includegraphics[width=15.9cm]{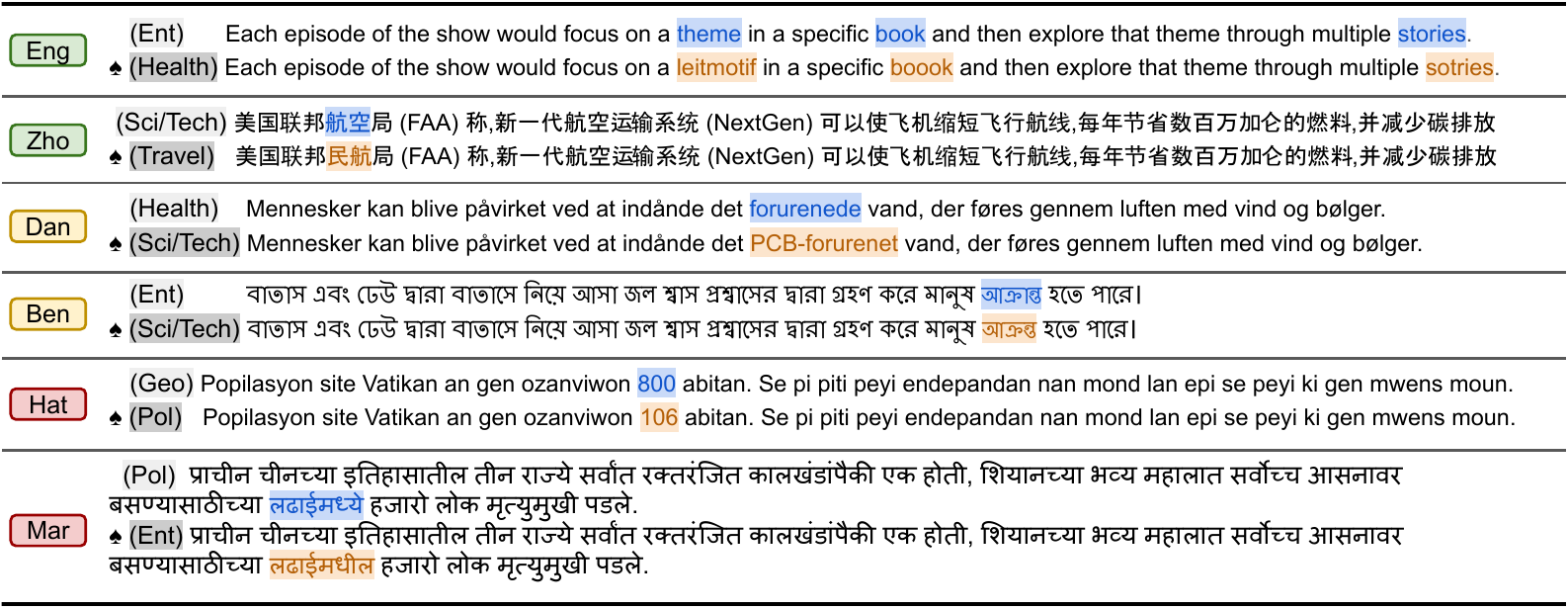}
    \caption{
    Examples of multilingual TextFooler applied to sentences from the SIB200 dataset. 
    For each pair, the original inputs are on top (true label in parentheses), and the adversarial input is below ($\spadesuit$ shows the flipped label from the attack). We highlight changes made from the original (blue) to the adversarial sample (orange). 
    Owing to a lack of resources like POS taggers for stricter quality control, we note the introduction of minor grammatical errors (see: Dan, the synonym has incorrect inflection), and changes to non-linguistic tokens (see: Hat) into adversarial samples. However, such samples remain helpful for diagnosing model security against adversarial attacks, especially as some MRLs and most LRLs lack spelling and grammar checkers. 
    }
    \label{fig:example}
\end{figure*}

\section{Methodology}\label{sec:methodology}
Consider a dataset of $n$ natural language inputs $X=\{x_1, x_2,\dots, x_n\}$ with corresponding labels $Y=\{y_1, y_2,\dots, y_n\}$. 
Given a finetuned classifier $F$, the goal of a black-box adversarial attack is to make minor perturbations to each $x_i$ sentence (\textit{i.e.}, $x_i \rightarrow x_i^* $) such that $F(x_i) \neq F(x_i^*)$. Importantly, the adversarial sentence $x_i^*$ should retain a near identical syntactic and semantic equivalence and fluency with the original $x_i$, to avoid suspicion.  
In this work, we leverage two main adversarial attack methods for generating $X^*$ in a multilingual context, described below. 

\paragraph{Multilingual TextFooler.}
TextFooler \cite{jin2020bertreallyrobuststrong} is a popular black-box adversarial attack for English, included as a baseline across many works in NLP Security. 
Given an input sequence $x$ comprised of $m$ words $x=\{w_1,w_2,...,w_m\}$, TextFooler generates an adversarial $x^*$ through a number of simple steps:

\begin{enumerate}[noitemsep,topsep=1pt]
    \item Excluding stop words, words are ranked by importance score $I$, determined by the change to the prediction as each $w_j$ is deleted. If a word's deletion does not alter the predicted label, then
    $I = P(y_{true}|x) - P(y_{true}|x_{delete\_w_j})$. If word deletion flips the label,
    $I = (P(y_{true}|x) - P(y_{true}|x_{delete\_w_j})) + (P(y_{flip}|x_{delete\_w_j}) - (P(y_{flip}|x) )$. 
    \item In order of their importance, words are assessed for replacement by synonyms. To be a eligible, $w_j$ must have candidate synonyms $c$ available in a pre-trained word embedding model, where the cosine similarity Cos($\mathrm{Emb}_{w_j}$, $\mathrm{Emb}_c$) $\geq$ similarity threshold $\delta$, and those synonyms must have the same POS tag as $w_j$.   
    \item The original word $w_j$ then is replaced by each remaining candidate synonyms $c$ as in $x^* = \{w_1, \dots, w_{j-1}, c ,w_{j+1}, \dots, w_{m}\}$, and passed through the classifier $F(x^*)$ to obtain the new prediction scores. 
    \item Among all adversarial samples $x^*$ which successfully flip the classifier's label, only the one with the highest similarity to the original is kept, 
    as determined by Cos($\mathrm{Emb}_x$, $\mathrm{Emb}_{x^*}$) using a sentence encoder. 
\end{enumerate}

Notably, several steps for English TextFooler ostensibly preclude MRLs and LRLs. 
Step 1 assumes the existence of stop word lists, unavailable for the majority of languages; accordingly, we do not check words against stop word lists in the importance score. 
At Step 2, original TextFooler leverages an English-only word embedding model by \citet{mrksic-etal-2016-counter}, which is specifically designed for synonym-lookup. As such, for each word $w_j$, they pull the top 50 synonyms, and set the similarity threshold $\delta=0.7$. To adopt this to a multilingual setting, we instead rely on FastText \cite{bojanowski2016enriching}, as its available in 157 languages, including our full set of 70. FastText was not specifically trained for synonym retrieval, however, making synonym retrieval slightly more challenging. We similarly pull the top 50 synonyms, and slightly lower $\delta$ to 0.6 cosine similarity threshold. 
Beyond pre-trained synonym embeddings, Step 2 also assumes access to and a POS tagger, which many MRLs and most LRLs languages lack\footnote{We find that only 45/70 languages have readily available POS taggers through Trankit.}. 
Finally at Step 4, the original TextFooler assumes language coverage in a sentence encoder such as SBERT \cite{reimers-2019-sentence-bert}. Unfortunately, sentence encoders are unavailable for the majority of the languages in this study, so we skip this step and keep only the first $x^*$ that flips the label. 

In summary, multilingual TextFooler works the same as the original, only without the additional quality control mechanisms. 
Examples generated with multilingual TextFooler can be seen in Table~\ref{fig:example}. 
Still, we end up with an average poisoning rate of 0.14 $\pm$ 0.07, similar to the original TextFooler.
Due to the lack of stop word lists, POS taggers, snynonym embeddings and sentence embedders for MRLs and LRLs, the stealthiness of our samples likely underperforms from that of English-only TextFooler. 
To verify that our adversarial samples are still true to the original label and are indeed viable for adversarial attacks, we include a human evaluation of the samples in Section~\ref{sec:results_analysis}.

\paragraph{Round-Trip Machine Translation Attack.}
As MT is the most common vehicle for bringing multilingualism into the fold of NLP Security, we also include an adversarial attack grounded in MT, inspired by \citet{yong2024lowresource}.
In their work, the authors implement a jailbreak attack (\textit{i.e.}, an adversarial attack which specifically targets LLM safety guardrails) whereby unsafe prompts in English are automatically translated to LRLs (\textit{e.g.}, Zulu), passed through GPT4, and the output is subsequently translated back to English, to determine whether the jailbreak succeeded. 
Although not directly comparable, we similarly use this ``round-trip'' approach, by translating the input language to Zulu, then back to the original language, before passing it to the classifier. The result is a slightly corrupted version of the original input, which serves as the adversarial sample.
For the MT model, we leverage NLLB-200-distilled-1.3B model\footnote{\url{https://huggingface.co/facebook/nllb-200-distilled-1.3B}} \cite{nllbteam2022languageleftbehindscaling}, as it covers all of the languages included in this study. We chose Zulu as the target language for round-trip MT following \citet{yong2024lowresource}, and also because it is unrelated to the languages in this study; the only Niger-Congo language in our study is Yoruba, which comes from a different branch (Yoruboid vs Bantu). 

\section{Experiments}
To evaluate the security of monolingual and multilingual LMs for LRLs, MRLs, and HRLs, we first finetune classifiers over monolingual data. The resulting classifiers are used to assess the efficacy of the adversarial attacks. 

\paragraph{Datasets.} 
To train the classifiers, we leverage two massively multilingual text classification datasets: \textbf{Taxi1500} \cite{ma2023taxi1500} and \textbf{SIB200} \cite{adelani-etal-2024-sib} (see Table~\ref{tab:datasets}). 
Both of these datasets have train, dev, and test sets for all 70 languages in our setup. 
We note that both of Taxi1500 and SIB200 are markedly smaller than commonly leveraged datasets for English-only adversarial attacks. For example, the Yelp, IMDB, and AG's News datasets have larger train and test sets by orders of magnitude (\textit{i.e}, 560K--38K, 25K--25K, and 120K--7.6K train--dev samples, respectively) and much longer text samples (\textit{i.e}, 152, 215, and 43 average words per sample), making English-only adversarial attacks significantly easier to develop and test than for other languages. Additionally, both Taxi1500 and SIB200 are imbalanced datasets; Taxi1500's most common class appears 4$\times$ more often than the least common class, and 3$\times$ for SIB200.

\begin{table}[t]
\centering
\small
\begin{tabular}{lcc}
\toprule
 & \multicolumn{1}{l}{\textbf{Taxi1500}} & \multicolumn{1}{l}{\textbf{SIB200}} \\
 \midrule
\textbf{Domain} & Bible & WikiMedia \\
\textbf{Langs} & 1502 & 205 \\
\textbf{Labels} & 6 & 7 \\
\textbf{Avg. Words} & 28 & 21 \\
\midrule
\textbf{Train} & 860 & 701 \\
\textbf{Dev} & 106 & 99 \\
\textbf{Test} & 111 & 204 \\ 
\bottomrule
\end{tabular}
\caption{The two topic classification datasets used in this study. 
For Taxi1500, Bibles for 823/1502 languages are publicly available, with the remainder sourced from \citet{mayer-cysouw-2014-creating}. 
SIB200 is derived from the ``DEV'' and ``DEVTEST'' sets of Flores-200, which is sourced from WikiNews, WikiJunior, and WikiVoyage. 
We report the average number of words per sample for English.
Both datasets employ 80/10/10 splits.   
}
\label{tab:datasets}
\end{table}

\paragraph{Victim Language Models.}
For all 70 languages in both datasets, we finetune classifiers over both monolingual and multilingual encoder-only LMs (see Table~\ref{tab:models} for overview).
For \textbf{monolingual} LMs, we leverage Goldfish \cite{chang2024goldfishmonolinguallanguagemodels}, a suite of transformer-based LMs. 
Despite their small size, the authors find that the Goldfish models outperform large, massively multilingual LMs like BLOOM and MaLA-500 for many LRLs. 
Goldfish models are also available for different amounts of training data (\textit{i.e.}, 5MB, 10MB, 100MB, 1GB), allowing for extensive examination of monolingual LMs in this study. 
The pretraining data for the Goldfish models covers a variety domains such as books and social media, however, according to the authors the domains are inconsistent across languages; for quality control, languages with only Bible data are dropped.

\begin{table}[t]
\centering
\small
\begin{tabular}{llccc}
\toprule
\textbf{Type} & \textbf{Model} & \multicolumn{1}{l}{\textbf{Base}} & \multicolumn{1}{l}{\textbf{\#L}} & \textbf{\#P} \\ \hline
\multirow{4}{*}{\textbf{Mono}} & goldfish-5mb & \multirow{4}{*}{GPT2} & 1 & 39M \\ %
 & goldfish-10mb &  & 1 & 39M \\ %
 & goldfish-100mb &  & 1 & 125M \\ %
 & goldfish-1000mb &  & 1 & 125M \\ %
 \midrule
\multirow{4}{*}{\textbf{Multi}} & mBERT & BERT & 101 & 197M \\
 & XLM-R base & RoBERTa & 100 & 279M \\
  & XLM-R large & RoBERTa & 100 & 560M \\
  & Glot500 & RoBERTa & 511 & 395M \\ 
 \bottomrule
\end{tabular}
\caption{The monolingual and multilingual LMs assessed in this work. Although these models have varied architectures and numbers of languages (\#L), the largest monolingual models are comparable in size to the smallest of the multilingual models in total number of parameters (\#P). 
The Goldfish models have a vocabulary size of 50k, mBERT 110k, XLMR-R 250k, and Glot500 401k. 
}
\label{tab:models}
\end{table}

For \textbf{multilingual} LMs, we use mBERT, XLM-R, and Glot500 \cite{devlin-etal-2019-bert, conneau-etal-2020-unsupervised, imani-etal-2023-glot500}.
Although encoder-only models like mBERT and XLMR are no longer the state-of-the-art, these LMs are still heavily relied upon in the realm of low-resource NLP (\textit{e.g}, \citet{adelani-etal-2024-comparing, goswami-etal-2025-multilingual, sani-etal-2025-investigating}), and thus remain relevant for study. Moreover, the ever-increasing dataset sizes in the newer and larger LMs have been demonstrated to \textit{hurt} performance for LRLs and MRLs \cite{chang-etal-2024-multilinguality}, further motivating a dedicated investigation of encoder-only LMs. 
In terms of domain, mBERT is pre-trained on Wikipedia, XLM-R on CCNet (open domain), and Glot500 on an amalgamation of 150 different datasets of varying domains. 

Finally, we note that most -- but not all -- of our 
70 languages are covered equally across our selection of LMs. For example, Amharic has all 4 Goldfish models and is seen by XLM-R and Glot500, but is unseen by mBERT. 
In the main body of this work, we present only the intersection of languages which have been seen by all 8 monolingual and multilingual models (n=45). 
The full results, where we include \textit{e.g.} Amharic in all calculations \textit{except for mBERT} can be found in Appendix~\ref{app:full}. We find the same trends carry across both the intersection and full set of languages.

\begin{table*}[t]
\begin{tabular}{lllll|lll}
\hline
 &  & \multicolumn{3}{c|}{\textbf{Taxi1500 (ASR)}} & \multicolumn{3}{c}{\textbf{SIB200 (ASR)}} \\
\multicolumn{1}{c}{\textbf{Attack}} & \multicolumn{1}{c}{\textbf{Model}} & \multicolumn{1}{c}{\textbf{LRLs}} & \multicolumn{1}{c}{\textbf{MRLs}} & \multicolumn{1}{c|}{\textbf{HRLs}} & \multicolumn{1}{c}{\textbf{LRLs}} & \multicolumn{1}{c}{\textbf{MRLs}} & \multicolumn{1}{c}{\textbf{HRLs}} \\ \hline
\multirow{8}{*}{TextFooler} & goldfish5mb & 0.84 $\pm$ 0.12 & 0.85 $\pm$ 0.13 & 0.82 $\pm$ 0.09 & 0.84 $\pm$ 0.10 & 0.90 $\pm$ 0.08 & 0.77 $\pm$ 0.22 \\
 & goldfish10mb & 0.77 $\pm$ 0.15 & 0.84 $\pm$ 0.11 & 0.77 $\pm$ 0.09 & 0.84 $\pm$ 0.09 & 0.85 $\pm$ 0.08 & 0.75 $\pm$ 0.22 \\
 & goldfish100mb & \textbf{0.62 $\pm$ 0.18} & \textbf{0.73 $\pm$ 0.10} & \textbf{0.70 $\pm$ 0.15} & 0.56 $\pm$ 0.07 & 0.54 $\pm$ 0.09 & 0.43 $\pm$ 0.19 \\
 & goldfish1000mb & 0.65 $\pm$ 0.16 & \textbf{0.73 $\pm$ 0.10} & \textbf{0.70 $\pm$ 0.16} & \textbf{0.50 $\pm$ 0.10} & \textbf{0.45 $\pm$ 0.07} & \textbf{0.34 $\pm$ 0.17} \\ \cline{2-8} 
 & mbert & 0.63 $\pm$ 0.11 & 0.72 $\pm$ 0.12 & 0.68 $\pm$ 0.09 & 0.57 $\pm$ 0.06 & 0.50 $\pm$ 0.13 & 0.45 $\pm$ 0.08 \\
 & xlmr-base & 0.54 $\pm$ 0.15 & 0.62 $\pm$ 0.12 & 0.56 $\pm$ 0.14 & 0.44 $\pm$ 0.09 & 0.41 $\pm$ 0.05 & 0.29 $\pm$ 0.14 \\
 & xlmr-large & 0.58 $\pm$ 0.15 & 0.64 $\pm$ 0.11 & 0.58 $\pm$ 0.14 & \textbf{0.39 $\pm$ 0.08} & \textbf{0.34 $\pm$ 0.05} & \textbf{0.26 $\pm$ 0.11} \\
 & glot500 & \textbf{0.52 $\pm$ 0.14} & \textbf{0.58 $\pm$ 0.14} & \textbf{0.54 $\pm$ 0.12} & 0.42 $\pm$ 0.09 & 0.45 $\pm$ 0.10 & 0.33 $\pm$ 0.15 \\ \hline
\multirow{8}{*}{RT-MT} & goldfish5mb & 0.38 $\pm$ 0.09 & 0.40 $\pm$ 0.13 & 0.39 $\pm$ 0.13 & 0.37 $\pm$ 0.09 & 0.36 $\pm$ 0.09 & 0.33 $\pm$ 0.09 \\
 & goldfish10mb & 0.33 $\pm$ 0.11 & 0.35 $\pm$ 0.13 & 0.34 $\pm$ 0.12 & 0.33 $\pm$ 0.07 & 0.32 $\pm$ 0.08 & 0.30 $\pm$ 0.05 \\
 & goldfish100mb & 0.23 $\pm$ 0.07 & 0.27 $\pm$ 0.11 & 0.25 $\pm$ 0.09 & 0.18 $\pm$ 0.03 & 0.19 $\pm$ 0.05 & 0.14 $\pm$ 0.03 \\
 & goldfish1000mb & \textbf{0.21 $\pm$ 0.05} & \textbf{0.26 $\pm$ 0.11} & \textbf{0.23 $\pm$ 0.09} & \textbf{0.15 $\pm$ 0.02} & \textbf{0.17 $\pm$ 0.06} & \textbf{0.12 $\pm$ 0.02} \\ \cline{2-8} 
 & mbert & 0.32 $\pm$ 0.09 & 0.27 $\pm$ 0.11 & 0.24 $\pm$ 0.11 & 0.34 $\pm$ 0.15 & 0.23 $\pm$ 0.11 & 0.14 $\pm$ 0.07 \\
 & xlmr-base & 0.20 $\pm$ 0.06 & 0.21 $\pm$ 0.09 & 0.17 $\pm$ 0.06 & 0.14 $\pm$ 0.02 & 0.15 $\pm$ 0.05 & \textbf{0.11 $\pm$ 0.02} \\
 & xlmr-large & 0.20 $\pm$ 0.05 & 0.23 $\pm$ 0.11 & 0.18 $\pm$ 0.08 & \textbf{0.13 $\pm$ 0.02} & 0.15 $\pm$ 0.05 & \textbf{0.11 $\pm$ 0.03} \\
 & glot500 & \textbf{0.19 $\pm$ 0.09} & \textbf{0.18 $\pm$ 0.08} & \textbf{0.14 $\pm$ 0.07} & 0.14 $\pm$ 0.04 & \textbf{0.14 $\pm$ 0.05} & \textbf{0.11 $\pm$ 0.02} \\ \hline
\end{tabular}
\caption{ASR scores for the set of languages with perfectly even model coverage (n=45 in total with LRLs=8, MRLs=30, HRLs=7). We bold the \textbf{lowest} ASR scores, which represent the \textbf{best-case scenario} \textit{i.e.}, where models are the most secure against adversarial attacks, in both monolingual and multilingual settings.}
\label{tab:asr_intersection}
\end{table*}

\paragraph{Finetuning, Attacking, and Metrics.}
Models are finetuned on the train set of Taxi1500 and SIB200, respectively. 
In Appendix~\ref{app:full} Table~\ref{tab:clean_acc_intersection}, we report the average clean accuracies (\textbf{CACC}) -- the accuracy over the regular datasets before any attack -- across LRLs, MRLs, and HRLs for the dev and test sets.
As LRLs and MRLs are reasonably expected to be less accurate than HRLs, we verify that all classifiers sufficiently beat the random weighted guessing score for each dataset\footnote{We also conducted experiments with mT5 (base and large), however, not all resulting classifiers surpassed the random weighted guessing baseline even after thorough hyperparameter tuning, and thus mt5 results are excluded from this work. For larger LMs, classifiers trained \textit{via} parameter-efficient finetuning are more appropriate. But as this setting cannot be directly compared against the smaller, monolingual models, we leave this for future work.}, where $k$ is the number of classes, and $p_i$ is the probability of a given class in $p_1, p_2, ...,p_k$, and  
random guessing accuracy is equal to $\sum_{i=1}^{k} p_i^2$.
All of the classifiers comfortably surpass this threshold, ensuring fair comparison across models, despite an expected performance divide between HRLs, MRLs and LRLs. 

For the adversarial attacks, we take all samples that a given model correctly predicted across both the dev and test sets (\textit{i.e.}, the union of dev and test samples where CACC is 100\%). 
The number of adversarial samples varies by language and model, but on average is 134 $\pm$ 52 samples. 
While this number is smaller than most studies in English-only adversarial attacks, we maintain strong statistical power through the number of datasets, languages, LMs, and random seeds in this study (resulting in over 3000 classifiers for evaluation). 
To evaluate the efficacy of the adversarial attacks, we use the attack success rate (\textbf{ASR}). As the previous CACC is 100\%, ASR is simply $1-$ the model's new accuracy over the adversarial samples.  

\paragraph{Implementation and Hardware Details.}
Each classifier is trained on a single AMD MI250x GPU, accessed through an HPC cluster. We finetune models for 15 epochs with early stopping. For all models, we use a batch size of 32, with an initial learning rate of 2e-5 and the AdamW optimizer. These settings were found to work well across the full suite of models during initial hyperparameter-tuning. For each resulting classifier, we save only the best model with regards to loss on the dev set. As training data sizes are not large, finetuning a single classifier is quick. We run all experiments across 3 random seeds, and report the averaged results. 

\section{Results and Analysis}\label{sec:results_analysis}

Results for multilingual TextFooler and Round-Trip MT are in Table~\ref{tab:asr_intersection} (and full results in Appendix~\ref{app:full} Table~\ref{tab:asr_full}).
We see that ASR on TextFooler is markedly higher than Round-Trip MT.
Consequently, 
simple MT-based attacks may indeed risk underestimating the threat level for these languages, in the face of more targeted attacks.

\paragraph{Monolingual vs Multilingual.}
Across all settings, the smallest monolingual models (goldfish5mb and 10mb) are the most vulnerable to adversarial attacks.
However, the goldfish100mb and 1000mb models are more secure than mBERT despite being smaller in model size, but less secure than the other larger multilingual LMs.
This finding draws attention to two key variables for LM security: model \textbf{size} and \textbf{multilinguality}. 
If size were the only important factor, one would expect the largest model (XLM-R large) to always be the most secure, especially as XLM-R large consistently has the best CACC (see: App~\ref{app:full} Table~\ref{tab:clean_acc_intersection}). But in practice, we find that the Glot500-based classifiers can have similar or better (\textit{i.e.} lower) scores, especially for the LRLs, signaling some security benefit from multilinguality.

\begin{table}[t]
\centering
\small
\begin{tabular}{llccc}
\toprule
\textbf{Type} & \textbf{Model} & \multicolumn{1}{l}{\textbf{Base}} & \multicolumn{1}{l}{\textbf{\#L}} & \textbf{\#P} \\ \hline
\multirow{3}{*}{\textbf{Mod}} & IndicBERTv2 & BERT & 26 & 278M \\
 & SlavicBERT & BERT & 4 & 180M \\ 
 & CINO-base & RoBERTa & 8 & 190M \\
 \bottomrule
\end{tabular}
\caption{These are the moderately multilingual LMs examined in this work. Number of languages (\#L), and total number of parameters (\#P). 
}
\label{tab:moderate}
\end{table}

\paragraph{Does Multilinguality Guarantee Improved Security?}
To further explore the role of multilinguality in LM Security, we turn to ``moderately'' multilingual LMs -- \textit{i.e.} LMs pre-trained on a small but specific set of languages. We train and evaluate classifiers for seen languages with IndicBERT v2 \cite{doddapaneni-etal-2023-towards}, and SlavicBERT \cite{arkhipov-etal-2019-tuning}, and CINO \cite{yang-etal-2022-cino} (Table~\ref{tab:moderate}). SlavicBERT and CINO have similar size to the monolingual goldfish100mb and 1000mb models, and all 3 LMs are smaller than Glot500 and XLM-R.

\begin{figure*}[t]
    \centering %
    \includegraphics[width=13cm, keepaspectratio]{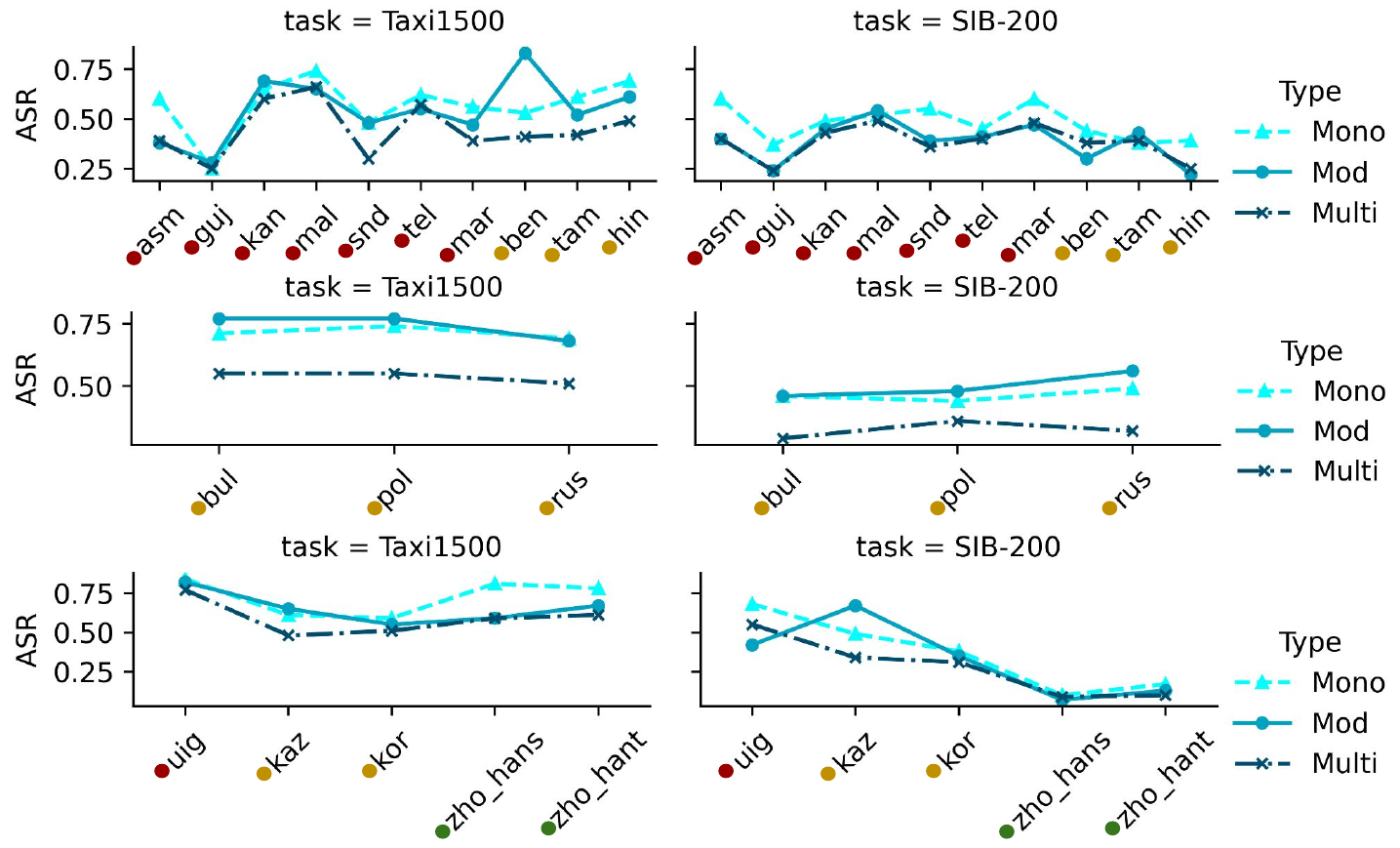}
    \caption{
   We compare the best Mono(lingual) and Multi(lingual) models to the relevant Mod(erately) multilingual model. On top the moderately multilingual LM is IndicBERT v2, middle is SlavicBERT, and bottom is CINO. 
   The dots by each language code indicate the level of resourcefulness (LRL=red, MRL=yellow, HRL=green). 
   Across all settings, we can see that the regular multilingual LMs are often the most secure (\textit{i.e.} having the lowest ASR).  
    }
    \label{fig:moderate}
\end{figure*}

In Figure~\ref{fig:moderate} we show the ASR for the moderately multilingual models, compared against the best performing monolingual and multilingual models for each language. 
We observe that moderately multilingual models do not consistently outperform the similarly-sized monolingual models, and results vary from language to language\footnote{We note that CINO is pretrained on kaz\_ara while Taxi1500 and SIB200 both pertain to kaz\_cyrl and thus the worsened security is expected here. In all other cases, the scripts are the same.}. This finding suggests that increasing multilinguality alone cannot guarantee improved security. 
Meanwhile, the moderately multilingual models seldom outperform the larger, standard multilingual LMs, though in some cases do perform on par. 
Both model size and multilinguality remain important factors for securing LMs for MRLs and LRLs.

\paragraph{Human Evaluation.}
Due to the dearth of resources for MRLs and LRLs, multilingual TextFooler necessarily lacks some of the quality control mechanisms of the original.  
To check the validity of the adversarial samples, we conduct a manual evaluation for 8 languages included in this study, stratified across the resource levels (\textit{i.e.}, eng, zho\_hans, hin, hun, ben, dan, hat, nor). For each language, we ask a native or highly proficient speaker to assess 10 randomly selected original-adversarial sentence pairs, in terms of the \textbf{type of change} (\textit{i.e.} typos, synonyms, synonyms+typos, other), \textbf{fluency and grammar} (\textit{i.e.}, fluent+grammatical, awkward, ungrammatical, awkward+ungrammatical), and \textbf{fidelity} to the original label (\textit{i.e.}, does the label of the adversarial sample change? Yes or no). Per the latter question, all annotators are NLP practitioners, with a reasonable understanding of adversarial samples. 
Please see Appendix~\ref{appendix:human} for the full task description and instructions provided to annotators. The human evaluation is conducted only over samples from SIB200 data, as WikiMedia is a more general domain relevant to most speakers. We do not manually evaluate adversarial samples from Taxi1500 as the religious domain often contains heavily stylized and archaic language \cite{mielke-etal-2019-kind} and is not uniformly relevant to all cultures \cite{mager-etal-2023-ethical}. 

The combined annotator scores reveal that synonyms are the mostly commonly observed change at 36.25\%, followed by ``other'' at 32.5\%, typo at 25\%, and a combination of synonyms+typos at 6.25\%. Annotators report that the ``other'' category consists of linguistic noise, varying word forms (both acceptable and unacceptable), words with the wrong case or verb inflection, and antonyms. Despite some undesirable changes, 36.25\% of adversarial samples were both fluent and grammatical, 31.25\% were awkward, 16.25\% were ungrammatical, and 16.25\% were both awkward and ungrammatical. While the unavailability of tools like POS taggers demonstrably impacts the overall quality of the samples, we note that access to these tools still cannot guarantee complete fluency and grammaticality; the original TextFooler also produces some adversarial samples which are awkward for native speakers of English (\textit{cf} Table 6, example 2 in \citet{jin2020bertreallyrobuststrong}).
Despite these quality issues, however, annotator results show that \textbf{93.75\% adversarial samples are true to the original label}. For the remaining 5/80 samples, annotators mention undesirable linguistic noise and synonyms which can possibly change the topic (\textit{e.g.}, the change of ``people'' $\rightarrow$ ``tourists'' could make the adversarial sample belong to both the original Science label and the Travel label). 
Ultimately, the human evaluation reveals that multilingual TextFooler overall produces faithful adversarial samples to a high degree, in light of practical constraints of working with MRLs and LRLs. Moreover, as basic grammar and spell checkers are still lacking for many of these languages \cite{lent-etal-2022-creole}, these samples still remain useful and relevant for red-teaming LMs for MRLs and LRLs, even if less so for HRLs. 

\section{Conclusion}
While the largest monolingual LMs prove more secure than the smallest multilingual model, the advantages of monolingualism ultimately lose out to the increased model sizes of the larger multilingual models. 
As the the lack of resources for MRLs and LRLs prohibits the creation of larger monolingual models for most languages, it stands that monolingual LMs cannot be relied on to ensure the LM security for speakers and communities of MRLs and LRLs. 
With that, we return to the original, outstanding problem: multilingual LMs are only ever as secure as their weakest links, and this vulnerability carries with it potentially grave consequences for society at large. In other words, the current lack of equitable language technology for all amounts to a major security vulnerability for us all.

Going forward, increasing efforts to improve LM security for MRLs and LRLs should be a top priority for researchers in NLP Security and we encourage others to consider a wider variety of languages in their work. 
While considering typologically diverse language samples is receiving increased attention in NLP more broadly \citep{ploeger-etal-2024-typological, ploeger-etal-2024-framework}, it is perhaps even more critical that this is addressed in the context of NLP Security, where bad actors are nearly guaranteed to exploit the worst-case scenario. 
As LRLs represent one of the most serious security vulnerabilities to multilingual models today, studying this is not only important for affected speakers and communities, but critical for making more secure NLP for all, regardless of native language.
We hope that the adversarial attack methods presented in this work can serve as a useful starting point for the development of new LM defense methods.

\section*{Ethics Statement}

\paragraph{Intended Use and Harm Minimization.}
Following the recommendations by \citet{lent2025nlpsecurityethicswild} for more ethical NLP Security research, we engage with the ethical ramifications of this work during the methodology and experiment design. 
As many MRLs and LRLs are already situated in vulnerable contexts \cite{blasi-etal-2022-systematic}, works in NLP Security engaging with LRLs demand increase scrutiny to ensure that communities for LRLs and MRLs not harmed in the process of research.  
Accordingly, \textbf{this work does not set out to introduce new and aggressive attacks targeting MRLs and LRLs.} 
Rather, the goal of this work is to extend existing attacks, which have a demonstrated track record in the field. In doing so, we build a bridge between the large body of contemporary works in English-only NLP Security to LRLs and MRLs.
The intended use of this work is for research purposes only. 
We acknowledge that that there is potential for this work to be misused and that all works in NLP Security naturally entail some threat of dual-use accordingly. 
However as this work extends an well-known, older attack method, the potential for misuse of the methodology predates this work. Finally, upon official publication, we will release our code to help others more quickly start experimenting with more red-teaming and blue-teaming efforts, with the ultimate goal of improved LM security for LRLs and MRLs. 

\paragraph{Responsible Disclosure.}
Prior to publication, we attempted to contact the model owners to disclose our findings. 
However, the security vulnerabilities explored in this paper are already well-known and represent endemic issues to LMs, which cannot simply be patched, in accordance with traditional cyber-security protocol.  
As such, we do not attempt \textit{coordinated} responsible disclosure. We believe that this paper will be more valuable as a vehicle for public disclosure and raising awareness, particularly to communities and NLP practitioners concerned with MRLs and LRLs.  

\paragraph{Against Techno-Solutionism.}
This work is primarily concerned with the security deficits for MRLs and LRLs in contemporary LMs.
In the conclusion of this paper, we call for increased efforts to engage with more languages in NLP Security in order to ``improve'' security. However, it is important to stress that LM security for MRLs and LRLs can likely never be \textit{solved}, as LM security vulnerabilities are largely endemic to these systems and promising to fully solve them through technological means alone is irresponsible \cite{DBLP:journals/corr/abs-2009-13676}. 
Thus, efforts to pursue security from imperfect LMs will require reaching beyond purely technological solutions. For meaningful work in this space, it will be necessary to collaborate with experts from other fields \textit{e.g.} cybersecurity and HCI, and also to engage with the communities who are most negatively impacted by these technologies \cite{kogkalidis2024tablesnumbersnumbers}.

\section*{Acknowledgments}
HL was supported by the Carlsberg Foundation, under the \textit{Semper Ardens: Accelerate} program (project nr. CF210454). 
This work benefited greatly from conversations and feedback from others. 
In particular, we would like to thank Esther Ploeger for discussion and feedback throughout the course of this work.
We also thank Nicholas Walker, Mike Zhang, and Yiyi Chen for discussions and feedback. 
For help with assessing the quality and diagnosing issues with multilingual TextFooler, we extend our thanks to Xiaoyu Luo, Russa Biswas, Marcell Fekete, Nathanial Robinson, and Johannes Bjerva. 
Finally, we thank DeiC for allocating us computing
resources on the LUMI cluster (project nr. DeiC-AAU-S5-412301).

\bibliography{anthology1, anthology2, custom}

\appendix

\section{Reproducibility Checklist}

\paragraph{Main Checklist.}

\begin{itemize}[noitemsep,topsep=1pt]
    \item  Includes a conceptual outline and/or pseudocode description of AI methods introduced (\textbf{yes}/partial/no/NA)
    \item  Clearly delineates statements that are opinions, hypothesis, and speculation from objective facts and results (\textbf{yes}/no)
    \item  Provides well marked pedagogical references for less-familiar readers to gain background necessary to replicate the paper (\textbf{yes}/no)
\end{itemize}

\paragraph{Theory Checklist.}

\begin{itemize}[noitemsep,topsep=1pt]
    \item Does this paper make theoretical contributions? (yes/\textbf{no}) 
\end{itemize}

\paragraph{Data Checklist.}

\begin{itemize}[noitemsep,topsep=1pt]
    \item Does this paper rely on one or more datasets? (\textbf{yes}/no)
    \item A motivation is given for why the experiments are conducted on the selected datasets (\textbf{yes}/partial/no/NA)
    \item All novel datasets introduced in this paper are included in a data appendix. (yes/partial/no/\textbf{NA})
    \item All novel datasets introduced in this paper will be made publicly available upon publication of the paper with a license that allows free usage for research purposes. (yes/partial/no/\textbf{NA})
    \item All datasets drawn from the existing literature (potentially including authors’ own previously published work) are accompanied by appropriate citations. (\textbf{yes}/no/NA)
    \item All datasets drawn from the existing literature (potentially including authors’ own previously published work) are publicly available. (\textbf{yes}/partial/no/NA)
    \item All datasets that are not publicly available are described in detail, with explanation why publicly available alternatives are not scientifically satisfying. (yes/partial/no/\textbf{NA})
\end{itemize}

\paragraph{Experiments Checklist.}

\begin{itemize}[noitemsep,topsep=1pt]
    \item Does this paper include computational experiments? (\textbf{yes}/no)
    \item Any code required for pre-processing data is included in the appendix. (yes/partial/\textbf{no}). \textbf{$\rightarrow$ We will release all code on Github upon official publication.}
    \item All source code required for conducting and analyzing the experiments is included in a code appendix. (yes/partial/\textbf{no}). \textbf{$\rightarrow$ We will release all code on Github.}
    \item All source code required for conducting and analyzing the experiments will be made publicly available upon publication of the paper with a license that allows free usage for research purposes. (\textbf{yes}/partial/no)
    \item All source code implementing new methods have comments detailing the implementation, with references to the paper where each step comes from (\textbf{yes}/partial/no)
    \item If an algorithm depends on randomness, then the method used for setting seeds is described in a way sufficient to allow replication of results. (\textbf{yes}/partial/no/NA)
    \item This paper specifies the computing infrastructure used for running experiments (hardware and software), including GPU/CPU models; amount of memory; operating system; names and versions of relevant software libraries and frameworks. (\textbf{yes}/partial/no)
    \item This paper formally describes evaluation metrics used and explains the motivation for choosing these metrics. (\textbf{yes}/partial/no)
    \item This paper states the number of algorithm runs used to compute each reported result. (\textbf{yes}/no)
    \item Analysis of experiments goes beyond single-dimensional summaries of performance (e.g., average; median) to include measures of variation, confidence, or other distributional information. (\textbf{yes}/no)
    \item The significance of any improvement or decrease in performance is judged using appropriate statistical tests (e.g., Wilcoxon signed-rank). (yes/partial/no/\textbf{NA})
    \item This paper lists all final (hyper-)parameters used for each model/algorithm in the paper’s experiments. (yes/\textbf{partial}/no/NA) \textbf{$\rightarrow$ Full details will be available in the Github repository upon official publication; in cases where we don't state the hyperparameters, we use the default values.}
    \item This paper states the number and range of values tried per (hyper-) parameter during development of the paper, along with the criterion used for selecting the final parameter setting. (yes/\textbf{partial}/no/NA)

\end{itemize}

\section{Limitations}\label{app:limitations}

\paragraph{Quality Control and Scale.}

In this work, we have done our best to manually assess samples, to ensure that Multilingual TextFooler scales for for HRLs, MRLs, and LRLs, alike. 
In English, the use of external tools like POS taggers and sentence embedders, combined with the tremendous magnitude of data, ensure the feasibility of producing reasonably ``stealthy'' adversarial samples, entailing fluency and grammaticality when viewed by humans. 
However, many MRLs and most LRLs by definition lack these resources,   
including grammar and spell checkers, which are often taken for granted.
Accordingly, our efforts to scale TextFooler multilingually come at some cost to quality. In order to quantify this cost, we provided analysis manually performed by humans. Still, the large scale of this work makes it prohibitive to manually check samples from all languages across all models. We encourages others adopting our methodology to similarly verify the quality of some samples, and invite others to continue improving upon this work.

\paragraph{Consistency and Comparability.}
As individual languages differ in their level of resourcefulness, a limitation of this work is perfect comparability across models. In the main body of the work, we do our best to combat this by limiting results to only those 45/70 languages which are perfectly represented across all 8 models. Even so, the number of evaluation samples per language are not the same, and in the full results with all 70 languages, the number of languages are not uniformly the same either. Unfortunately, this inconvenience is the reality of working with MRLs and LRLs. 

Similarly for the models selected for study in this work, we compare a variety of model sizes and architectures. Due to practical constraints (\textit{e.g.} no BERT- or XLMR-based monolingual models), however, this paper lacks  a true "apples to apples" comparison of models. 
The models chosen in this work could also be argued to be ``outdated'', however, due to small datasizes, these encoder-only models are still heavily-used in lower-resourced NLP.
As such, this work presents a thorough comparison of the models as used in practice. For future work, larger LMs and different model finetuning techniques for classification should be investigated. 

\paragraph{Data Contamination.}
We do not check whether any of the LMs included in this study were already pretrained over samples included in the evaluation datasets. The Taxi1500 and SIB200 are Bible and WikiMedia domains, respectively; Goldfish and Glot500 pretraining data can include both Bible and Wikipedia domains for some languages. As XLM-R is trained on CCNet, these models were likely also exposed to these domains. Thus it is likely that there is some degree of data contamination across these experiments. Again, as resources are scarce for most languages, this is an expected limitation.

\section{Human Evaluation}\label{appendix:human}

\paragraph{Task Description.}
In this setup, we have taken "Original" input sentences, and made small changes to obtain an "Adversarial" sentences. 
These small changes can include typos or word substitution with similar words (we will call these "synonyms", although the scope here is more open than the strict definition of a synonym; it will be words which are used in a similar context as the original word). 
However, sometimes the small changes can be undesirable noise (see Example Row 29 of the annotation spreadsheet). 
Importantly, the Adversarial sentence should be close -- but not identical -- to the original, and retain the same semantic meaning as the original. Ideally, the adversarial sample is also fluent (\textit{i.e.}, reads like a natural sentence to a native speaker) and grammatical (\textit{i.e.} no grammar errors).	

You will be given 10 pairs of Original-Adversarial sentences.
For each Original sentence, there is also a True Topic label, designating the ground truth topic of the Original text. 
For each pair, in Column D, please select the \textbf{type of change observed} in the Adversarial sentence from the Original (\textit{i.e.} typo, synonym, synonym+typo, other).	
To help you, we have done our best to highlight the differences between original and adversarial in blue and orange, respectively -- but it is very possible that I have missed some differences. If you notices difference between the original and adversarial which are NOT highlighted, please still take them into consideration for your annotations. 	
Next, please assess the Adversarial sample for its \textbf{fluency and grammaticality} in Column E. A sample can be Fluent+Grammatical, simply Awkward (it reads unnaturally, but it is still technically grammatical), simply Ungrammatical (it is ungrammatical, but still reads mostly naturally), or both Awkward+Ungrammatical. 	
Lastly, in Column F, provide your judgment on whether the changes in the adversarial sample could lead a reasonably-trained classifier to deviate from the True Topic Label (\textit{i.e.}, \textbf{fidelity} to the original label). For example, if the general topic of the Adversarial sample remains unchanged and the changes are not too egregious, the answer would be "No". However, if the adversarial sample deviates too greatly from the Original (\textit{e.g.}, if there is a great deal of noise, or the substitutions have significantly shifted the meaning of the adversarial sample), the answer would be "Yes" (see Example Row 29 of the annotation spreadsheet). Please feel free to document any comments you may have. For example, if the type of change is "Other", perhaps you want to describe what has happened.

\section{Languages and Model Coverage}\label{appendix:resources}

Table~\ref{tab:languages_A} shows the HRLs and MRLs covered in this work, and Table~\ref{tab:languages_B} shows the LRLs. We note our sample is diverse in terms of language families and branches. Table~\ref{tab:models_A} details the model coverage of HRLs and MRLs and Table~\ref{tab:models_B} shows the LRLs. 

\begin{table*}[ht]
\begin{tabular}{lllllll}
\hline
\textbf{Cat.} & \textbf{Language} & \textbf{Family} & \textbf{Branch} & \textbf{Taxi1500} & \textbf{SIB200} & \textbf{Fasttext} \\ \hline
5 & arabic & Afro-Asiatic & Semitic & arb & arb\_Arab & ar \\
5 & chinese & Sino-tibetan & Sinitic & zho & zho\_Hans & zh \\
5 & chinese\_traditional & Sino-tibetan & Sinitic & zho & zho\_Hant & zh \\
5 & english & Indo-European & Germanic & eng & eng\_Latn & en \\
5 & french & Indo-European & Romance & fra & fra\_Latn & fr \\
5 & german & Indo-European & Germanic & deu & deu\_Latn & de \\
5 & japanese & Japonic & 0 & jpn & jpn\_Jpan & ja \\
5 & spanish & Indo-European & Romance & spa & spa\_Latn & es \\ \hline
4 & catalan & Indo-European & Romance & cat & cat\_Latn & ca \\
4 & croatian & Indo-European & Balto-Slavic & hrv & hrv\_Latn & hr \\
4 & finnish & Finnic-Uralic & 0 & fin & fin\_Latn & fi \\
4 & hindi & Indo-European & Indo-Iranian & hin & hin\_Deva & hi \\
4 & hungarian & Finnic-Uralic & 0 & hun & hun\_Latn & hu \\
4 & italian & Indo-European & Romance & ita & ita\_Latn & it \\
4 & korean & Koreanic & 0 & kor & kor\_Hang & ko \\
4 & polish & Indo-European & Balto-Slavic & pol & pol\_Latn & pl \\
4 & portuguese & Indo-European & Romance & por & por\_Latn & pt \\
4 & russian & Indo-European & Balto-Slavic & rus & rus\_Cyrl & ru \\
4 & serbian & Indo-European & Balto-Slavic & srp & srp\_Cyrl & sr \\
4 & swedish & Indo-European & Germanic & swe & swe\_Latn & sv \\
4 & turkish & Turkic & 0 & tur & tur\_Latn & tr \\
4 & vietnamese & Austroasiatic & Vietic & vie & vie\_Latn & vi \\
3 & afrikaans & Indo-European & Germanic & afr & afr\_Latn & af \\
3 & belarusian & Indo-European & Balto-Slavic & bel & bel\_Cyrl & be \\
3 & bengali & Indo-European & Indo-Iranian & ben & ben\_Beng & bn \\
3 & bulgarian & Indo-European & Balto-Slavic & bul & bul\_Cyrl & bg \\
3 & cebuano & Austronesian & 0 & ceb & ceb\_Latn & ceb \\
3 & danish & Indo-European & Germanic & dan & dan\_Latn & da \\
3 & egyptian\_arabic & Afro-Asiatic & Semitic & arz & arz\_Arab & arz \\
3 & estonian & Finnic-Uralic & 0 & est & est\_Latn & et \\
3 & hebrew & Afro-Asiatic & Semitic & heb & heb\_Hebr & he \\
3 & indonesian & Austronesian & 0 & ind & ind\_Latn & id \\
3 & kazakh & Turkic & 0 & kaz & kaz\_Cyrl & kk \\
3 & lithuanian & Indo-European & Balto-Slavic & lit & lit\_Latn & lt \\
3 & malay & Austronesian & 0 & msa & zsm\_Latn & ms \\
3 & slovenian & Indo-European & Balto-Slavic & slv & slv\_Latn & sl \\
3 & tagalog & Austronesian & 0 & tgl & tgl\_Latn & tl \\
3 & tamil & Dravidian & 0 & tam & tam\_Taml & ta \\
3 & thai & Kra-Dai & Tai & tha & tha\_Thai & th \\
3 & ukrainian & Indo-European & Balto-Slavic & ukr & ukr\_Cyrl & uk \\
3 & urdu & Indo-European & Indo-Iranian & urd & urd\_Arab & ur \\ \hline
\end{tabular}
\caption{The high and medium-resourced languages covered in this work. The category (Cat.) is derived from \citet{joshi-etal-2020-state}; we assign category 5 as HRLs, and categories 4-3 as MRLs, following previous work \cite{ustun2024ayamodelinstructionfinetuned}.}
\label{tab:languages_A}
\end{table*}

\begin{table*}[ht]
\begin{tabular}{lllllll}
\hline
\textbf{Cat.} & \textbf{Language} & \textbf{Family} & \textbf{Branch} & \textbf{Taxi1500} & \textbf{SIB200} & \textbf{Fasttext} \\ \hline
2 & amharic & Afro-Asiatic & Semitic & amh & amh\_Ethi & am \\
2 & haitian\_creole & Indo-European & Creole & hat & hat\_Latn & ht \\
2 & irish & Indo-European & Celtic & gle & gle\_Latn & ga \\
2 & marathi & Indo-European & Indo-Iranian & mar & mar\_Deva & mr \\
2 & yoruba & Niger-Congo & Atlantic-Congo & yor & yor\_Latn & yo \\
1 & assamese & Indo-European & Indo-Iranian & asm & asm\_Beng & as \\
1 & azerbaijani & Turkic & 0 & azb & azb\_Arab & azb \\
1 & bashkir & Turkic & 0 & bak & bak\_Cyrl & ba \\
1 & central\_kurdish & Indo-European & Indo-Iranian & ckb & ckb\_Arab & ckb \\
1 & esperanto & Constructed & 0 & epo & epo\_Latn & eo \\
1 & gujarati & Indo-European & Indo-Iranian & guj & guj\_Gujr & gu \\
1 & iloko & Austronesian & 0 & ilo & ilo\_Latn & ilo \\
1 & javanese & Austronesian & 0 & jav & jav\_Latn & jv \\
1 & kannada & Dravidian & 0 & kan & kan\_Knda & kn \\
1 & kirgyz & Turkic & 0 & kir & kir\_Cyrl & ky \\
1 & kurdish & Indo-European & Indo-Iranian & kmr & kmr\_Latn & ku \\
1 & maithili & Indo-European & Indo-Iranian & mai & mai\_Deva & mai \\
1 & malayalam & Dravidian & 0 & mal & mal\_Mlym & ml \\
1 & norwegian & Indo-European & Germanic & nob & nob\_Latn & no \\
1 & norwegian bokmål & Indo-European & Germanic & nob & nob\_Latn & no \\
1 & nynorsk & Indo-European & Germanic & nno & nno\_Latn & nn \\
1 & sindhi & Indo-European & Indo-Iranian & snd & snd\_Arab & sd \\
1 & sinhala & Indo-European & Indo-Iranian & sin & sin\_Sinh & si \\
1 & sudanese & Austronesian & 0 & sun & sun\_Latn & su \\
1 & tatar & Turkic & 0 & tat & tat\_Cyrl & tt \\
1 & telugu & Dravidian & 0 & tel & tel\_Telu & te \\
1 & turkmen & Turkic & 0 & tuk & tuk\_Latn & tk \\
1 & uyghur & Turkic & 0 & uig & uig\_Arab & ug \\
1 & waray & Austronesian & 0 & war & war\_Latn & war \\ \hline
\end{tabular}
\caption{The lower-resourced languages covered in this work. The category (Cat.) is derived from \citet{joshi-etal-2020-state}; we assign categories 2-1 as LRLs, following previous work \cite{ustun2024ayamodelinstructionfinetuned}. }
\label{tab:languages_B}
\end{table*}

\begin{table*}[ht]
\begin{tabular}{llllllll}
\hline
\textbf{Language} & \textbf{goldfish\_5mb} & \textbf{goldfish\_10mb} & \textbf{goldfish\_100mb} & \textbf{goldfish\_1000mb} & \textbf{mbert} & \textbf{xlmr} & \textbf{glot500} \\ \hline
arabic & arb\_arab & arb\_arab & arb\_arab & arb\_arab & ara & ara & arb\_Arab \\
chinese & zho\_hans & zho\_hans & zho\_hans & zho\_hans & zho\_hans & zho\_hans & zho\_Hani \\
chinese\_traditional & zho\_hant & zho\_hant & zho\_hant & 0 & zho\_hant & zho\_hant & zho\_Hani \\
english & eng\_latn & eng\_latn & eng\_latn & eng\_latn & eng & eng & eng\_Latn \\
french & fra\_latn & fra\_latn & fra\_latn & fra\_latn & fre & fre & fra\_Latn \\
german & deu\_latn & deu\_latn & deu\_latn & deu\_latn & ger & ger & deu\_Latn \\
japanese & jpn\_jpan & jpn\_jpan & jpn\_jpan & jpn\_jpan & jpn & jpn & jpn\_Jpan \\
spanish & spa\_latn & spa\_latn & spa\_latn & spa\_latn & spa & spa & spa\_Latn \\ \hline
catalan & cat\_latn & cat\_latn & cat\_latn & cat\_latn & cat & cat & cat\_Latn \\
croatian & hrv\_latn & hrv\_latn & hrv\_latn & hrv\_latn & hrv & hrv & hrv\_Latn \\
finnish & fin\_latn & fin\_latn & fin\_latn & fin\_latn & fin & fin & fin\_Latn \\
hindi & hin\_deva & hin\_deva & hin\_deva & hin\_deva & hin & hin & hin\_Deva \\
hungarian & hun\_latn & hun\_latn & hun\_latn & hun\_latn & hun & hun & hun\_Latn \\
italian & ita\_latn & ita\_latn & ita\_latn & ita\_latn & ita & ita & ita\_Latn \\
korean & kor\_hang & kor\_hang & kor\_hang & kor\_hang & kor & kor & kor\_Hang \\
polish & pol\_latn & pol\_latn & pol\_latn & pol\_latn & pol & pol & pol\_Latn \\
portuguese & por\_latn & por\_latn & por\_latn & por\_latn & por & por & por\_Latn \\
russian & rus\_cyrl & rus\_cyrl & rus\_cyrl & rus\_cyrl & rus & rus & rus\_Cyrl \\
serbian & srp\_cyrl & srp\_cyrl & srp\_cyrl & srp\_cyrl & srp & srp & srp\_Cyrl \\
swedish & swe\_latn & swe\_latn & swe\_latn & swe\_latn & swe & swe & swe\_Latn \\
turkish & tur\_latn & tur\_latn & tur\_latn & tur\_latn & tur & tur & tur\_Latn \\
vietnamese & vie\_latn & vie\_latn & vie\_latn & vie\_latn & vie & vie & vie\_Latn \\
afrikaans & afr\_latn & afr\_latn & afr\_latn & afr\_latn & afr & afr & afr\_Latn \\
belarusian & bel\_cyrl & bel\_cyrl & bel\_cyrl & bel\_cyrl & bel & bel & bel\_Cyrl \\
bengali & ben\_beng & ben\_beng & ben\_beng & ben\_beng & ben & ben & ben\_Beng \\
bulgarian & bul\_cyrl & bul\_cyrl & bul\_cyrl & bul\_cyrl & bul & bul & bul\_Cyrl \\
cebuano & ceb\_latn & ceb\_latn & ceb\_latn & 0 & ceb & 0 & ceb\_Latn \\
danish & dan\_latn & dan\_latn & dan\_latn & dan\_latn & dan & dan & dan\_Latn \\
egyptian\_arabic & arz\_arab & arz\_arab & arz\_arab & 0 & 0 & 0 & arz\_Arab \\
estonian & est\_latn & est\_latn & est\_latn & est\_latn & est & est & est\_Latn \\
hebrew & heb\_hebr & heb\_hebr & heb\_hebr & heb\_hebr & heb & heb & heb\_Hebr \\
indonesian & ind\_latn & ind\_latn & ind\_latn & ind\_latn & ind & ind & ind\_Latn \\
kazakh & kaz\_cyrl & kaz\_cyrl & kaz\_cyrl & kaz\_cyrl & kaz & kaz & kaz\_Cyrl \\
lithuanian & lit\_latn & lit\_latn & lit\_latn & lit\_latn & lit & lit & lit\_Latn \\
malay & msa\_latn & msa\_latn & msa\_latn & msa\_latn & msa & msa & msa\_Latn \\
slovenian & slv\_latn & slv\_latn & slv\_latn & slv\_latn & slv & slv & slv\_Latn \\
tagalog & tgl\_latn & tgl\_latn & tgl\_latn & tgl\_latn & tgl & 0 & tgl\_Latn \\
tamil & tam\_taml & tam\_taml & tam\_taml & tam\_taml & tam & tam & tam\_Taml \\
thai & tha\_thai & tha\_thai & tha\_thai & tha\_thai & tha & tha & tha\_Thai \\
ukrainian & ukr\_cyrl & ukr\_cyrl & ukr\_cyrl & ukr\_cyrl & ukr & ukr & ukr\_Cyrl \\
urdu & urd\_arab & urd\_arab & urd\_arab & urd\_arab & urd & urd & urd\_Arab \\ \hline
\end{tabular}
\caption{Model coverage for the HRLs (top) and MRLs (bottom) covered in this work. The 0 indicates that either no model exists for the given languages (\textit{e.g.}, there is no goldfish1000mb model for Cebuano) or that the language is \textit{unseen} in model pretraining (\textit{e.g.}, XLM-R has not been pre-trained on Tagalog).} 
\label{tab:models_A}
\end{table*}

\begin{table*}[h]
\begin{tabular}{llllllll}
\hline
\textbf{Language} & \textbf{goldfish\_5mb} & \textbf{goldfish\_10mb} & \textbf{goldfish\_100mb} & \textbf{goldfish\_1000mb} & \textbf{mbert} & \textbf{xlmr} & \textbf{glot500} \\ \hline
amharic & amh\_ethi & amh\_ethi & amh\_ethi & amh\_ethi & 0 & amh & amh\_Ethi \\
haitian\_creole & hat\_latn & hat\_latn & hat\_latn & 0 & hat & 0 & hat\_Latn \\
irish & gle\_latn & gle\_latn & gle\_latn & 0 & gle & gle & gle\_Latn \\
marathi & mar\_deva & mar\_deva & mar\_deva & mar\_deva & mar & mar & mar\_Deva \\
yoruba & yor\_latn & yor\_latn & yor\_latn & 0 & yor & 0 & yor\_Latn \\
assamese & asm\_beng & asm\_beng & asm\_beng & 0 & 0 & asm & asm\_Beng \\
azerbaijani & azb\_arab & azb\_arab & azb\_arab & 0 & azb & 0 & azb\_Arab \\
bashkir & bak\_cyrl & bak\_cyrl & bak\_cyrl & 0 & bak & 0 & bak\_Cyrl \\
central\_kurdish & ckb\_arab & ckb\_arab & ckb\_arab & 0 & 0 & 0 & ckb\_Arab \\
esperanto & epo\_latn & epo\_latn & epo\_latn & epo\_latn & 0 & epo & epo\_Latn \\
gujarati & guj\_gujr & guj\_gujr & guj\_gujr & guj\_gujr & guj & guj & guj\_Gujr \\
iloko & ilo\_latn & ilo\_latn & 0 & 0 & 0 & 0 & ilo\_Latn \\
javanese & jav\_latn & jav\_latn & jav\_latn & 0 & jav & jav & jav\_Latn \\
kannada & kan\_knda & kan\_knda & kan\_knda & kan\_knda & kan & kan & kan\_Knda \\
kirgyz & kir\_cyrl & kir\_cyrl & kir\_cyrl & kir\_cyrl & kir & kir & kir\_Cyrl \\
kurdish & kmr\_latn & kmr\_latn & 0 & 0 & 0 & 0 & kmr\_Latn \\
maithili & mai\_deva & mai\_deva & 0 & 0 & 0 & 0 & mai\_Deva \\
malayalam & mal\_mlym & mal\_mlym & mal\_mlym & mal\_mlym & mal & mal & mal\_Mlym \\
norwegian & nor\_latn & nor\_latn & nor\_latn & nor\_latn & nob & nob & nor\_Latn \\
norwegian bokmål & nob\_latn & nob\_latn & nob\_latn & nob\_latn & nob & nob & nob\_Latn \\
nynorsk & nno\_latn & nno\_latn & nno\_latn & 0 & nno & 0 & nno\_Latn \\
sindhi & snd\_arab & snd\_arab & snd\_arab & 0 & 0 & snd & snd\_Arab \\
sinhala & sin\_sinh & sin\_sinh & sin\_sinh & sin\_sinh & 0 & sin & sin\_Sinh \\
sudanese & sun\_latn & sun\_latn & sun\_latn & 0 & sun & sun & sun\_Latn \\
tatar & tat\_cyrl & tat\_cyrl & tat\_cyrl & tat\_cyrl & tat & 0 & tat\_Cyrl \\
telugu & tel\_telu & tel\_telu & tel\_telu & tel\_telu & tel & tel & tel\_Telu \\
turkmen & tuk\_latn & tuk\_latn & tuk\_latn & 0 & 0 & 0 & tuk\_Latn \\
uyghur & uig\_arab & uig\_arab & uig\_arab & 0 & 0 & uig & uig\_Arab \\
waray & war\_latn & war\_latn & war\_latn & 0 & war & 0 & war\_Latn \\ \hline
\end{tabular}
\caption{Model coverage for the LRLs covered in this work. The 0 indicates that either no model exists for the given languages (\textit{e.g.}, there is no goldfish1000mb model for Haitian Creole) or that the language is \textit{unseen} in model pretraining (\textit{e.g.}, mBERT has not been pre-trained on Amharic).}
\label{tab:models_B}
\end{table*}

\section{Full Results}\label{app:full}
Table~\ref{tab:clean_acc_intersection} contains the CACC for the 45 languages included in the main body of the paper, which had complete and even coverage for all models. 
Table~\ref{tab:clean_acc_FULL} contains the CACC for the full 70 languages, excluding unseen languages. 
Finally, Table~\ref{tab:asr_full} contains the full ASR scores for all 70 languages, excluding unseen languages.

\begin{table*}[]
\begin{tabular}{lllll|lll}
\hline
\multicolumn{1}{c}{\textbf{}} & \multicolumn{1}{c}{\textbf{}} & \multicolumn{3}{c|}{\textbf{Taxi1500 (CACC)}} & \multicolumn{3}{c}{\textbf{SIB200 (CACC)}} \\
\textbf{Split} & \textbf{Model} & \multicolumn{1}{c}{\textbf{LRLs}} & \multicolumn{1}{c}{\textbf{MRLs}} & \multicolumn{1}{c|}{\textbf{HRLs}} & \multicolumn{1}{c}{\textbf{LRLs}} & \multicolumn{1}{c}{\textbf{MRLs}} & \multicolumn{1}{c}{\textbf{HRLs}} \\ \hline
\multirow{8}{*}{Val Acc} & goldfish5mb & 0.53 $\pm$ 0.07 & 0.48 $\pm$ 0.07 & 0.47 $\pm$ 0.06 & 0.41 $\pm$ 0.09 & 0.43 $\pm$ 0.12 & 0.43 $\pm$ 0.10 \\
 & goldfish10mb & 0.52 $\pm$ 0.06 & 0.50 $\pm$ 0.06 & 0.52 $\pm$ 0.07 & 0.49 $\pm$ 0.09 & 0.52 $\pm$ 0.13 & 0.51 $\pm$ 0.11 \\
 & goldfish100mb & 0.61 $\pm$ 0.03 & 0.60 $\pm$ 0.06 & 0.59 $\pm$ 0.05 & 0.81 $\pm$ 0.04 & 0.84 $\pm$ 0.04 & 0.86 $\pm$ 0.03 \\
 & goldfish1000mb & \textbf{0.66 $\pm$ 0.03} & \textbf{0.64 $\pm$ 0.05} & \textbf{0.62 $\pm$ 0.07} & \textbf{0.86 $\pm$ 0.02} & \textbf{0.87 $\pm$ 0.03} & \textbf{0.89 $\pm$ 0.02} \\ \cline{2-8} 
 & mbert & 0.58 $\pm$ 0.03 & 0.58 $\pm$ 0.05 & 0.60 $\pm$ 0.06 & 0.82 $\pm$ 0.06 & 0.84 $\pm$ 0.09 & 0.88 $\pm$ 0.03 \\
 & xlmr-base & 0.68 $\pm$ 0.04 & 0.67 $\pm$ 0.05 & 0.64 $\pm$ 0.06 & 0.86 $\pm$ 0.03 & 0.87 $\pm$ 0.02 & 0.88 $\pm$ 0.02 \\
 & xlmr-large & \textbf{0.74 $\pm$ 0.04} & \textbf{0.72 $\pm$ 0.08} & \textbf{0.73 $\pm$ 0.05} & \textbf{0.88 $\pm$ 0.02} & \textbf{0.90 $\pm$ 0.02} & \textbf{0.90 $\pm$ 0.02} \\
 & glot500 & 0.63 $\pm$ 0.07 & 0.60 $\pm$ 0.09 & 0.61 $\pm$ 0.12 & 0.80 $\pm$ 0.12 & 0.78 $\pm$ 0.17 & 0.83 $\pm$ 0.13 \\ \hline
\multirow{8}{*}{Test Acc} & goldfish5mb & 0.47 $\pm$ 0.04 & 0.50 $\pm$ 0.05 & 0.47 $\pm$ 0.04 & 0.35 $\pm$ 0.06 & 0.41 $\pm$ 0.10 & 0.42 $\pm$ 0.07 \\
 & goldfish10mb & 0.50 $\pm$ 0.05 & 0.52 $\pm$ 0.06 & 0.53 $\pm$ 0.05 & 0.44 $\pm$ 0.05 & 0.50 $\pm$ 0.10 & 0.54 $\pm$ 0.06 \\
 & goldfish100mb & 0.61 $\pm$ 0.05 & 0.61 $\pm$ 0.06 & 0.61 $\pm$ 0.05 & 0.82 $\pm$ 0.03 & 0.83 $\pm$ 0.03 & 0.85 $\pm$ 0.02 \\
 & goldfish1000mb & \textbf{0.67 $\pm$ 0.05} & \textbf{0.66 $\pm$ 0.06} & \textbf{0.65 $\pm$ 0.05} & \textbf{0.86 $\pm$ 0.02} & \textbf{0.88 $\pm$ 0.02} & \textbf{0.89 $\pm$ 0.02} \\ \cline{2-8} 
 & mbert & 0.59 $\pm$ 0.03 & 0.60 $\pm$ 0.05 & 0.63 $\pm$ 0.03 & 0.78 $\pm$ 0.05 & 0.82 $\pm$ 0.11 & 0.86 $\pm$ 0.03 \\
 & xlmr-base & 0.67 $\pm$ 0.04 & 0.66 $\pm$ 0.05 & 0.65 $\pm$ 0.05 & 0.85 $\pm$ 0.02 & 0.88 $\pm$ 0.02 & 0.89 $\pm$ 0.01 \\
 & xlmr-large & \textbf{0.71 $\pm$ 0.04} & \textbf{0.70 $\pm$ 0.09} & \textbf{0.71 $\pm$ 0.05} & \textbf{0.88 $\pm$ 0.02} & \textbf{0.89 $\pm$ 0.02} & \textbf{0.89 $\pm$ 0.02} \\
 & glot500 & 0.61 $\pm$ 0.06 & 0.59 $\pm$ 0.08 & 0.58 $\pm$ 0.11 & 0.79 $\pm$ 0.13 & 0.75 $\pm$ 0.18 & 0.80 $\pm$ 0.14 \\ \hline
\end{tabular}
\caption{Clean acc (CACC) over the set of languages with perfectly even model coverage  (\textit{i.e.}, from our original 70 languages, only \textbf{n=45} have perfect coverage across all 8 models, resulting in LRLs=8, MRLs=30, HRLs=7). Here, the \textbf{random weighted guess} would be 19.56\% Acc. for Taxi1500 and 16.51\% Acc fo SIB200; accordingly, all of our models comfortably beat this baseline. Here we bold the best scores. Across the board, we see that goldfish1000mb and XLM-R large have the best CACC.}
\label{tab:clean_acc_intersection}
\end{table*}

\begin{table*}[]
\begin{tabular}{lllll|lll}
\hline
 &  & \multicolumn{3}{c|}{\textbf{Taxi1500 (CACC)}} & \multicolumn{3}{c}{\textbf{SIB200 (CACC)}} \\
\textbf{Split} & \textbf{Model} & \multicolumn{1}{c}{\textbf{LRLs}} & \multicolumn{1}{c}{\textbf{MRLs}} & \multicolumn{1}{c|}{\textbf{HRLs}} & \multicolumn{1}{c}{\textbf{LRLs}} & \multicolumn{1}{c}{\textbf{MRLs}} & \multicolumn{1}{c}{\textbf{HRLs}} \\ \hline
\multirow{8}{*}{Val Acc} & goldfish5mb & 0.51 $\pm$ 0.06 & 0.48 $\pm$ 0.07 & 0.48 $\pm$ 0.06 & 0.42 $\pm$ 0.10 & 0.43 $\pm$ 0.12 & 0.42 $\pm$ 0.10 \\
 & goldfish10mb & 0.53 $\pm$ 0.06 & 0.50 $\pm$ 0.06 & 0.52 $\pm$ 0.07 & 0.50 $\pm$ 0.10 & 0.51 $\pm$ 0.13 & 0.51 $\pm$ 0.11 \\
 & goldfish100mb & 0.60 $\pm$ 0.07 & 0.59 $\pm$ 0.07 & 0.59 $\pm$ 0.05 & 0.79 $\pm$ 0.07 & 0.83 $\pm$ 0.04 & 0.87 $\pm$ 0.03 \\
 & goldfish1000mb & \textbf{0.65 $\pm$ 0.07} & \textbf{0.64 $\pm$ 0.05} & \textbf{0.62 $\pm$ 0.07} & \textbf{0.86 $\pm$ 0.02} & \textbf{0.87 $\pm$ 0.03} & \textbf{0.89 $\pm$ 0.02} \\ \cline{2-8} 
 & mbert & 0.57 $\pm$ 0.04 & 0.58 $\pm$ 0.05 & 0.60 $\pm$ 0.06 & 0.79 $\pm$ 0.08 & 0.84 $\pm$ 0.09 & 0.89 $\pm$ 0.03 \\
 & xlmr-base & 0.59 $\pm$ 0.11 & 0.66 $\pm$ 0.06 & 0.65 $\pm$ 0.07 & 0.76 $\pm$ 0.17 & 0.87 $\pm$ 0.02 & 0.89 $\pm$ 0.02 \\
 & xlmr-large & \textbf{0.65 $\pm$ 0.12} & \textbf{0.71 $\pm$ 0.09} & \textbf{0.74 $\pm$ 0.05} & \textbf{0.81 $\pm$ 0.14} & \textbf{0.90 $\pm$ 0.02} & \textbf{0.90 $\pm$ 0.02} \\
 & glot500 & 0.60 $\pm$ 0.10 & 0.60 $\pm$ 0.09 & 0.61 $\pm$ 0.11 & 0.73 $\pm$ 0.18 & 0.78 $\pm$ 0.17 & 0.84 $\pm$ 0.12 \\ \hline
\multirow{8}{*}{Test Acc} & goldfish5mb & 0.49 $\pm$ 0.06 & 0.49 $\pm$ 0.05 & 0.48 $\pm$ 0.05 & 0.39 $\pm$ 0.07 & 0.40 $\pm$ 0.10 & 0.42 $\pm$ 0.08 \\
 & goldfish10mb & 0.51 $\pm$ 0.06 & 0.52 $\pm$ 0.07 & 0.54 $\pm$ 0.05 & 0.46 $\pm$ 0.06 & 0.49 $\pm$ 0.10 & 0.54 $\pm$ 0.06 \\
 & goldfish100mb & 0.60 $\pm$ 0.07 & 0.61 $\pm$ 0.07 & 0.61 $\pm$ 0.05 & 0.79 $\pm$ 0.06 & 0.83 $\pm$ 0.04 & 0.85 $\pm$ 0.02 \\
 & goldfish1000mb & \textbf{0.67 $\pm$ 0.07} & \textbf{0.66 $\pm$ 0.06} & \textbf{0.65 $\pm$ 0.05} & \textbf{0.86 $\pm$ 0.02} & \textbf{0.88 $\pm$ 0.02} & \textbf{0.89 $\pm$ 0.02} \\ \cline{2-8} 
 & mbert & 0.56 $\pm$ 0.05 & 0.59 $\pm$ 0.05 & 0.63 $\pm$ 0.03 & 0.76 $\pm$ 0.08 & 0.82 $\pm$ 0.11 & 0.86 $\pm$ 0.03 \\
 & xlmr-base & 0.58 $\pm$ 0.10 & 0.65 $\pm$ 0.06 & 0.65 $\pm$ 0.05 & 0.74 $\pm$ 0.17 & 0.87 $\pm$ 0.03 & 0.89 $\pm$ 0.01 \\
 & xlmr-large & \textbf{0.63 $\pm$ 0.11} & \textbf{0.69 $\pm$ 0.09} & \textbf{0.72 $\pm$ 0.05} & \textbf{0.79 $\pm$ 0.14} & \textbf{0.89 $\pm$ 0.03} & \textbf{0.89 $\pm$ 0.02} \\
 & glot500 & 0.57 $\pm$ 0.09 & 0.58 $\pm$ 0.08 & 0.59 $\pm$ 0.11 & 0.71 $\pm$ 0.19 & 0.75 $\pm$ 0.18 & 0.81 $\pm$ 0.13 \\ \hline
\end{tabular}
\caption{Clean Accuracy (CACC) for the full set of \textbf{n=70, for seen languages only} (LRLs=29, MRLs=33, HRLs=29). For example, Amharic has all 4 goldfish models and is seen by all of the multilingual models \textit{except for mBERT}; this table thus includes the CACC for Amharic in the average scores for LRLs for every model \textit{except for mBERT}. The \textbf{random weighted guess} would be 19.56\% Acc. for Taxi1500 and 16.51\% Acc fo SIB200; accordingly, all of our models comfortably beat this baseline. Here we bold the best scores. Across the board, we see that goldfish1000mb and XLM-R large have the best CACC.}
\label{tab:clean_acc_FULL}
\end{table*}

\begin{table*}[]
\begin{tabular}{lllll|lll}
\hline
 &  & \multicolumn{3}{c|}{\textbf{Taxi1500 (ASR)}} & \multicolumn{3}{c}{\textbf{SIB200 (ASR)}} \\
\multicolumn{1}{l}{\textbf{Attack}} & \multicolumn{1}{l}{\textbf{Model}} & \multicolumn{1}{c}{\textbf{LRLs}} & \multicolumn{1}{c}{\textbf{MRLs}} & \multicolumn{1}{c|}{\textbf{HRLs}} & \multicolumn{1}{c}{\textbf{LRLs}} & \multicolumn{1}{c}{\textbf{MRLs}} & \multicolumn{1}{c}{\textbf{HRLs}} \\ \hline
\multirow{8}{*}{TextFooler} & goldfish5mb & 0.78 $\pm$ 0.16 & 0.85 $\pm$ 0.12 & 0.84 $\pm$ 0.09 & 0.85 $\pm$ 0.10 & 0.89 $\pm$ 0.08 & 0.75 $\pm$ 0.21 \\
 & goldfish10mb & 0.75 $\pm$ 0.14 & 0.84 $\pm$ 0.12 & 0.79 $\pm$ 0.10 & 0.85 $\pm$ 0.09 & 0.85 $\pm$ 0.08 & 0.71 $\pm$ 0.22 \\
 & goldfish100mb & \textbf{0.66 $\pm$ 0.16} & 0.74 $\pm$ 0.10 & 0.71 $\pm$ 0.14 & 0.64 $\pm$ 0.14 & 0.55 $\pm$ 0.10 & 0.40 $\pm$ 0.20 \\
 & goldfish1000mb & 0.68 $\pm$ 0.15 & \textbf{0.73 $\pm$ 0.10} & \textbf{0.70 $\pm$ 0.16} & \textbf{0.52 $\pm$ 0.09} & \textbf{0.45 $\pm$ 0.07} & \textbf{0.34 $\pm$ 0.17} \\ \cline{2-8} 
 & mbert & 0.66 $\pm$ 0.13 & 0.72 $\pm$ 0.12 & 0.69 $\pm$ 0.08 & 0.61 $\pm$ 0.14 & 0.51 $\pm$ 0.13 & 0.44 $\pm$ 0.08 \\
 & xlmr-base & 0.60 $\pm$ 0.15 & 0.62 $\pm$ 0.12 & 0.57 $\pm$ 0.13 & 0.56 $\pm$ 0.18 & 0.42 $\pm$ 0.08 & 0.27 $\pm$ 0.14 \\
 & xlmr-large & 0.62 $\pm$ 0.16 & 0.64 $\pm$ 0.12 & 0.59 $\pm$ 0.14 & 0.53 $\pm$ 0.19 & \textbf{0.35 $\pm$ 0.08} & \textbf{0.24 $\pm$ 0.12} \\
 & glot500 & \textbf{0.53 $\pm$ 0.16} & \textbf{0.59 $\pm$ 0.14} & \textbf{0.55 $\pm$ 0.11} & \textbf{0.50 $\pm$ 0.15} & 0.45 $\pm$ 0.11 & 0.31 $\pm$ 0.15 \\ \hline
\multirow{8}{*}{RT-MT} & goldfish5mb & 0.41 $\pm$ 0.14 & 0.39 $\pm$ 0.13 & 0.41 $\pm$ 0.15 & 0.41 $\pm$ 0.09 & 0.36 $\pm$ 0.09 & 0.33 $\pm$ 0.08 \\
 & goldfish10mb & 0.38 $\pm$ 0.15 & 0.34 $\pm$ 0.13 & 0.36 $\pm$ 0.13 & 0.35 $\pm$ 0.09 & 0.32 $\pm$ 0.08 & 0.29 $\pm$ 0.04 \\
 & goldfish100mb & 0.31 $\pm$ 0.18 & 0.27 $\pm$ 0.11 & 0.27 $\pm$ 0.10 & 0.23 $\pm$ 0.09 & 0.19 $\pm$ 0.05 & 0.14 $\pm$ 0.03 \\
 & goldfish1000mb & \textbf{0.25 $\pm$ 0.14} & \textbf{0.26 $\pm$ 0.11} & \textbf{0.23 $\pm$ 0.09} & \textbf{0.15 $\pm$ 0.02} & \textbf{0.17 $\pm$ 0.06} & 0.12 $\pm$ 0.02 \\ \cline{2-8} 
 & mbert & 0.35 $\pm$ 0.13 & 0.27 $\pm$ 0.11 & 0.25 $\pm$ 0.10 & 0.30 $\pm$ 0.14 & 0.22 $\pm$ 0.11 & 0.13 $\pm$ 0.06 \\
 & xlmr-base & 0.26 $\pm$ 0.13 & 0.21 $\pm$ 0.09 & 0.18 $\pm$ 0.07 & 0.19 $\pm$ 0.10 & 0.16 $\pm$ 0.05 & 0.12 $\pm$ 0.03 \\
 & xlmr-large & 0.27 $\pm$ 0.14 & 0.22 $\pm$ 0.11 & 0.19 $\pm$ 0.08 & 0.19 $\pm$ 0.09 & 0.15 $\pm$ 0.05 & \textbf{0.11 $\pm$ 0.03} \\
 & glot500 & \textbf{0.21 $\pm$ 0.10} & \textbf{0.18 $\pm$ 0.07} & \textbf{0.15 $\pm$ 0.07} & \textbf{0.16 $\pm$ 0.08} & \textbf{0.14 $\pm$ 0.05} & \textbf{0.11 $\pm$ 0.03} \\ \hline
\end{tabular}
\caption{TextFooler ASR results for the full set of \textbf{n=70, for seen languages only} (LRLs=29, MRLs=33, HRLs=29). For example, Amharic has all 4 goldfish models and is seen by all of the multilingual models \textit{except for mBERT}; this table thus includes the ASR's for Amharic in the average scores for LRLs for every model \textit{except for mBERT}. We bold the \textbf{lowest} ASR scores, which represent the \textbf{best-case scenario} \textit{i.e.}, where models are the most secure against adversarial attacks, in both monolingual and multilingual settings.}
\label{tab:asr_full}
\end{table*}

\end{document}